\documentclass[12pt,twoside]{article}
\usepackage{graphicx,url}
\usepackage[british]{babel} \usepackage{amsmath}
\usepackage{amssymb}
\usepackage[tight]{subfigure}
\usepackage{latexsym}
\usepackage{hyperref}  \usepackage{algorithm2e}
\usepackage{amsthm}

\newdimen\paravsp  \paravsp=1.3ex
\newcommand{\beq}{\begin{equation}}    \newcommand{\eeq}{\end{equation}}
\newcommand{\beqn}{\begin{displaymath}}\newcommand{\eeqn}{\end{displaymath}}
\newcommand{\bqa}{\begin{eqnarray}}    \newcommand{\eqa}{\end{eqnarray}}
\newcommand{\bqan}{\begin{eqnarray*}}  \newcommand{\eqan}{\end{eqnarray*}}
\newcommand{\paradot}[1]{\vspace{\paravsp plus 0.5\paravsp minus 0.5\paravsp}\noindent{\bf\boldmath{#1.}}}
\newcommand{\paranodot}[1]{\vspace{\paravsp plus 0.5\paravsp minus 0.5\paravsp}\noindent{\bf\boldmath{#1}}}
\newcommand{\req}[1]{(\ref{#1})}
\newcommand{\eps}{\varepsilon}

\newcommand{\Euro}{$\,$C$\!\!\!\!\!$\raisebox{0.2ex}{=}}

\newcommand{\SetR}{I\!\!R}
\newcommand{\SetN}{I\!\!N}

\newcommand{\SetZ}{Z\!\!\!Z}

\newcommand{\E}{{\bf E}}
\renewcommand{\P}{\text{Pr}}
\renewcommand{\v}{\boldsymbol}
\newcommand{\trp}{{\!\top\!}}

\newcommand{\Loss}{\mbox{Loss}}
\newcommand{\tr}{\mbox{tr}}
\newcommand{\X}{\v \theta}
\newcommand{\Xortho}{\v \theta_{ortho}}
\newcommand{\Xpersp}{\v \theta_{persp}}
\renewcommand{\sec}{{Section}}
\newcommand{\fig}{{Figure}}

\begin{document}
\title{\vspace{-4ex}
\vskip 2mm\bf\Large\hrule height5pt \vskip 4mm
A Novel Illumination-Invariant Loss for Monocular 3D Pose Estimation
\vskip 4mm \hrule height2pt}
\author{{\bf Srimal Jayawardena} and {\bf Marcus Hutter} and {\bf Nathan Brewer}\\[3mm]
\normalsize Research School of Computer Science\\[-0.5ex]
\normalsize Australian National University \\[-0.5ex]
\normalsize Canberra, ACT, 0200, Australia \\
\normalsize \texttt{\{srimal.jayawardena, di.yang, marcus.hutter\}@anu.edu.au}
}
\maketitle
\maketitle

\begin{abstract}
The problem of identifying the 3D pose of a known
object from a given 2D image has important applications in
Computer Vision. Our proposed method of registering a 3D
model of a known object on a given 2D photo of the object has
numerous advantages over existing methods. It does not require
prior training, knowledge of the camera parameters, explicit
point correspondences or matching features between the image
and model. Unlike techniques that estimate a partial 3D pose (as
in an overhead view of traffic or machine parts on a conveyor
belt), our method estimates the complete 3D pose of the object. It
works on a single static image from a given view under varying
and unknown lighting conditions. For this purpose we derive
a novel illumination-invariant distance measure between the 2D
photo and projected 3D model, which is then minimised to find
the best pose parameters. Results for vehicle pose detection in
real photographs are presented.

\end{abstract}

%------------------------------------------------------------
\section{Introduction}\label{secIntro}
%------------------------------------------------------------
\begin{figure}[h!]
\centering
\def\w{0.48}
\def\tp{T269_SANY2383_800x600_pngTweak_detect_ellipse_gc20_best_}
\subfigure[Initial rough pose]{\label{fig:ResultsAstinaBestRough}
	 \includegraphics[width=\w\textwidth]{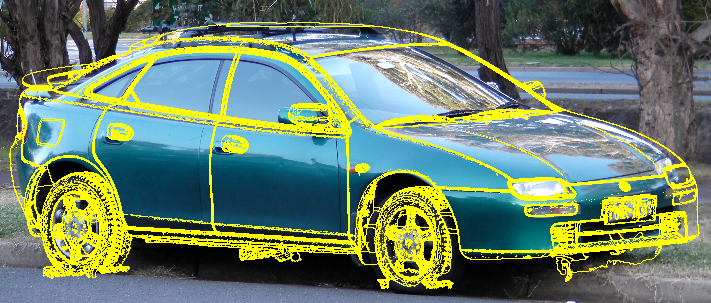}}
\subfigure[Final pose]{\label{fig:ResultsAstinaBestFine}
	 \includegraphics[width=\w\textwidth]{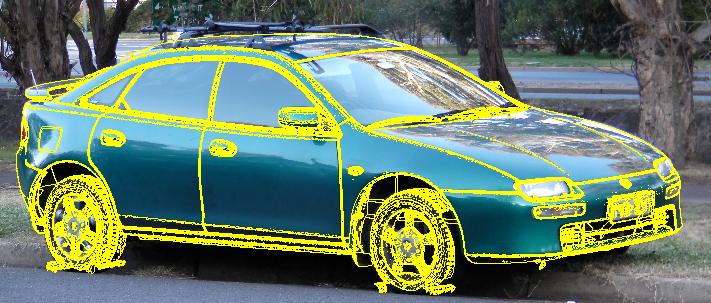}}
\caption[Results]{Recovered pose of a Mazda Astina using a scanned 3D model of the car.  \ref{fig:ResultsAstinaBestRough} shows the `Initial rough pose' (from the wheel match method) used to initialize the optimization. \ref{fig:ResultsAstinaBestFine} shows the resulting  `Final pose' (a perfect match) obtained by optimizing the novel loss function. The pose is shown in `yellow' by an outline of the projected 3D model. The images have been cropped for visual clarity. Note the large amount of reflection in the front of the car, which make pose recovery very challenging with conventional methods.}
\label{fig:ResultsAstinaBest}
\end{figure}
Pose estimation is a fundamental problem in computer vision
and has applications in robotic vision and intelligent image analysis.
In general, pose estimation refers to the process of
obtaining the location and orientation of an object and its parts.
We restrict our work to non-articulated objects where there is no relative movement between object parts.
The accuracy and nature of the pose estimate required varies from application to application.
Certain applications require the
estimation of the full 3D pose of an object, while other
applications require only a subset of the pose parameters.

\paradot{Motivation}
The 2D-3D registration problem in particular is concerned with
estimating the pose parameters that describe a 3D object model
within a given 2D scene. An image/photograph of a known object can
be analyzed in greater detail if a 3D CAD model of the object can be
registered over it (as in {\fig} \ref{fig:ResultsAstinaBestFine}) to be used as a ground truth.
A target application is automatic damage detection in vehicles using photos taken by a non-expert.
The photos will be taken in an uncontrolled environment (where the orientation of the vehicle and camera parameters are unknown) and delivered to a server for analysis. We restrict ourselves to cases where the vehicle is not completely destroyed.
The focus of this work is to develop a method to estimate the pose
of a known 3D object model in a given 2D image, with an emphasis on
estimating the pose of vehicles. We have the following objectives in
mind.
\begin{itemize}
\item Use only a single, static image limited to a single view
\item Work with any unknown camera (without prior camera calibration)
\item Avoid user interaction
\item Avoid prior training / learning
\item Work under varying and unknown lighting conditions
\item Estimate the full 3D pose of the object (not a partial
    pose as in an overhead view of traffic or machine parts
    along a conveyor belt)
\item Work in an uncontrolled environment
\end{itemize}
A 3D pose estimation method with these properties would also be useful in remote sensing, automated scene recognition and computer graphics, as it allows for additional information to be extracted without the need for human involvement.

Existing pose estimation methods include point correspondence based \cite{david2004softposit}\cite{moreno2008pose}, implicit
shape model based \cite{implicitshapepose} and image gradient based \cite{3dposegrayval}\cite{tan2000efficient} methods. However, these
methods do not fully satisfy the objectives
mentioned above, hence the necessity of our novel method. A detailed
review of existing pose estimation methods ranging over the past 30
years is presented in Section~\ref{secRelatedWork}.

\paradot{Main contribution}
This paper presents a method which registers a known 3D model onto a
given 2D photo containing the modeled object while satisfying the
objectives outlined above.
It does this by measuring the closeness of the projected 3D model to
the 2D photo on a pixel (rather than feature) basis. Background and
unknown lighting conditions of the photo are major complications,
which prevent using a naive image difference like the absolute or
square loss as a measure of fit.

The major contribution of this paper is the novel ``distance'' measure
in Section~\ref{secMatching3Dto2D} that does neither depend on the
lighting of the real scene in the photo nor on choosing an
appropriate lighting in the rendering of the 3D model, and hence does
not require knowledge of the lighting. Technically, we derive in
Section~\ref{secLossFunction} a loss function for vector-valued
pixel attributes (of different modality) that is invariant under
linear transformations of the attributes.

The loss functions is analyzed using synthetic and real photographs in Section \ref{secLossLandscape}.
We show that the loss function is well behaved and can be optimized using a standard optimization method to find an accurate pose.
Optimizing the loss function is described in Section \ref{secOptim}.
Sensitivity of the final recovered pose to the initialization and results on real photographs are presented in Section \ref{secResults}.
Implementation details are discussed in Section \ref{secTech}.

%------------------------------------------------------------
\section{Related Work}\label{secRelatedWork}
%------------------------------------------------------------
Model based object recognition has received considerable attention
in the computer vision community.
A survey  by Chin and Dyer
\cite{modelbasedsurvey1986} shows that model based object
recognition  algorithms generally fall into three categories based on
the  type of object representation used - namely 2D representations,
2.5D representations and 3D representations.

\paranodot{2D}
representations store the information of a particular 2D view of
an object (a characteristic view) as a model and use this
information to identify the object from a 2D image. Global feature
methods have been used by Gleason and Algin \cite{gleason1979} to
identify objects like spanners and nuts on a conveyor belt. Such
methods use features such as the area, perimeter, number of holes
visible and other global features to model the object. Structural
features like boundary segments have been
used by Perkins \cite{perkins1978} to detect machine parts using 2D
models. A relational graph method has been used by Yachida and Tsuji
\cite{yachida1977} to match objects to a 2D model using graph
matching techniques.
These 2D representation based algorithms require prior training of
the system using a `show by example' method.

\paranodot{2.5D}
approaches are also viewer centered, where the object is known to
occur in  a particular view. They differ from the 2D approach as the
model stores additional information such as intrinsic image
parameters and surface-orientation maps. The work done by Poje  and
Delp \cite{poje1982} explain the use of intrinsic scene parameters
in the form of range (depth) maps and needle (local surface
orientation) maps. Shape from shading \cite{horn1975} and
photometric stereo \cite{woodham1978photometric} are some other
examples of the use of the 2.5D approach used for the recognition of
industrial parts.
A range of techniques for such 2D/2.5D representations are described
by Forsythe and Ponce \cite{forsythe}, by posing the object
recognition problem as a correspondence problem. These methods
obtain a hypothesis based on the correspondences of a few matching
points in the image and the model. The hypothesis is validated
against the remaining known points.

\paranodot{3D}
approaches are utilized in situations where the object of
interest can appear in a scene from multiple viewing angles. Common
3D representation approaches can be either an `exact representation'
or a `multi-view feature representation'. The latter method uses a
composite model  consisting of  2D/2.5D models for a limited set of
views. Multi-view feature representation is used along with the
concept of generalized cylinders  by Brooks and Binford
\cite{brookes1981} to detect different types of industrial motors in
the so  called ACRONYM system. The models used in the exact
representation method, on the contrary, contain an exact
representation of the complete 3D object. Hence a 2D projection of
the object can be created for any desired  view. Unfortunately, this
method is often considered too costly in terms of processing time.

\paradot{Limitations}
The 2D and 2.5D representations are insufficient for general
purpose applications. For example, in the case of vehicle damage detection, a vehicle may be
photographed from an arbitrary view in order to indicate the damaged parts.
Similarly, the 3D multi-view feature representation is unsuitable
as it restricts the pose of the object to a limited set of views.
Therefore, an exact 3D representation is preferred.
Little work has been done to date on
identifying the pose of an exact 3D model from a single 2D image.
Huttenlocher and Ullman \cite{huttenlocher1990recognizing}  use
a 3D model that contains the locations of edges. The
edges/contours identified in the 2D image are matched against
the edges in the 3D model to calculate the pose of the object.
 The method has been implemented for simple 3D objects.
However, it is unclear if this method will work well on objects with rounded surfaces
without clearly identifiable edges.

\paradot{Image gradients}
Gray scale image gradients have been used to estimate the 3D pose in
traffic  video footage from a stationary camera by Kollnig and Nagel
\cite{3dposegrayval}.  The method compares image gradients  instead
of simple edge segments, for better performance. Image gradients
from projected polyhedral models are compared against image
gradients in video images. 
The pose is formulated using 3 degrees of freedom; 2 for position
and 1 for angular orientation. Tan and Baker \cite{tan2000efficient}
use image gradients and a Hough transform based algorithm for
estimating  vehicle pose in traffic scenes, once more describing the
pose via 3 degrees of freedom. Pose estimation using 3 degrees of
freedom  is adequate for traffic image sequences, where the camera
position remains fixed with respect to the ground plane.
However, this approach does not provide a full pose estimate
required for a general purpose application.

\paradot{Implicit Shape Models}
Recent work  by Arie-Nachimson and Ronen Basri
\cite{implicitshapepose} makes use of `implicit shape models' to
recognize 3D objects from 2D images.
The model consists of a set of
learned features, their 3D locations and the views in which they are
visible.
The learning process is further refined using factorization
methods. The pose estimation consists of evaluating the
transformations of the features that give the best match. 
A typical model requires around 65 images to be trained.
Many different vehicle models exist and new ones are manufactured frequently.
Hence, methods that require training vehicle models are too laborious and time consuming for our work.

\paranodot{Feature-based methods}
\cite{david2004softposit,moreno2008pose} attempt to simultaneously solve  the pose and
point correspondence problems. The success of these methods are
affected by the quality of the features extracted from the object.
Objects like vehicles have large homogeneous regions which yield very sparse features.
Also, the highly reflective surfaces in vehicles generate a lot of false positives. Our method on the contrary, does not depend on feature extraction.

\paranodot{Distance metrics}
can be used to represent a distance between two data sets, and hence
give a measure of their similarity. Therefore, distance metrics can
be used to measure similarity between different 2D images, as well
as 2D images and 2D projections of a 3D model. A basic distance metric
would be the \emph{Euclidean Distance} or the 2-norm $||\cdot||_2$.
However, this has the disadvantage of being dependent on the scale
of measurement.
We use the \emph{Mahalanobis Distance} \cite{mahalanobis1936generalized} for our work, which
is a scale-invariant distance measure.
It is used by Xing et al.\ \cite{xing2003distance} for
clustering. It is also used by Deriche and Faugeras
\cite{deriche1990tracking} to match line segments in a sequence of
time varying images.

%-------------------------------------------------------------------------------------------------------------------
\section{Matching 3D Models and 2D Photos  using the Invariant Loss}\label{secMatching3Dto2D}
%-------------------------------------------------------------------------------------------------------------------
We describe our approach of matching 3D models to 2D photos in this
section using a novel illumination-invariant loss function. A
detailed derivation of the loss is provided in
Section~\ref{secLossFunction}.

\paradot{The problem}
Assume we want to match a 3D model ($M$) to a 2D photo ($F$) or vice
versa. More precisely, we have a 3D model (e.g.\ as a triangulated
textured surface) and we want to find a projection $\theta$ for
which the rendered 2D image $M_\theta$ has the same perspective as
the 2D photo $F$. As long as we do not know the lighting conditions
of $F$, we cannot expect $F$ to be close to $M_\theta$, even for the
correct $\theta$. Indeed, if the light in $F$ came from the right,
but the light shines on $M$ from the left, $M_\theta$ may be close to
the negative of $F$.

\paradot{Setup}
Formally, let $P=\SetZ_{n_x}\times
\SetZ_{n_y}=\{1,...,n_x\}\times\{1,...,n_y\}$ be the set of
$|P|$ (integer) pixel coordinates, and $p=(x,y)\in P$ be a pixel coordinate.
Let $F:P\to\SetR^n$ be a photo with $n$ real pixel attributes, and
$M_\theta:P\to\SetR^m$ be a projection of a 3D object using pose parameters $\X$ to a 2D image
with $m$ real pixel attributes.
Possible attributes include colours, local
texture features or surface normals.
In the following we
consider the case of gray-level photos ($n=1$), and for reasons that
will become clear, use surface normals and brightness $(m=4)$ of the
(projected) 3D model.

\paradot{Lambertian reflection model}
A simple Lambertian reflection model \cite{Foley:95} is not realistic enough to
result in a zero loss on real photos, even at the correct pose.
Nevertheless (we believe and experimentally confirm that) it
results in a minimum at the correct pose, which is sufficient
for matching purposes.

We use Phong shading without specular reflection for
this purpose \cite{Foley:95}. Let $I_a\in\SetR$ and $I_d\in\SetR$ be the global
ambient and diffuse light intensities of the 3D scene.
Let  $\v L\in\SetR^3$ be the (global) unit vector pointing towards the
light source (or their weighted sum in case of multiple sources).
For reasons to become clear later, we introduce an extra
illumination offset $I_0\in\SetR$ (which is 0 in the Phong model). For each
surface point $p$, let $k_a(p)\in\SetR$ and  $k_d(p)\in\SetR$ be the ambient
and diffuse reflection constants (intrinsic surface brightness) and
$\v\phi(p)\in\SetR^3$ be the unit (interpolated) surface ``normal''
vector. Then the apparent intensity $I$ of the corresponding point
$p$ in the projection $M_\theta(p)$ is
\beq
  I(p) = k_a(p) I_a \!+\! k_d(p)(\v L^\trp\v\phi(p))I_d \!+\! I_0
    \equiv A\!\cdot\!M_\theta(p) \!+\! b
\eeq
The last expression is the same as the first, just written in a
more convenient form:
$M_\theta(p):=(k_a(p),k_d(p)\v\phi(p))^\trp\in\SetR^{4\times 1}$ are the
known surface (dependent) parameters, and $A:=(I_a,I_d\v
L^\trp)\in\SetR^{1\times 4}$ are the four (unknown) global
illumination constants, and $b=I_0$. Since $I(\cdot)$ is linear
in $A$ and $M_\theta(\cdot)$, any rendering is a simple global
linear function of $M_\theta(p)$.
This model remains exact even for multiple light sources
and can easily be generalized to color models and color photos.

\paradot{Illumination invariant loss}
We measure the closeness of the projected 3D model $M_\theta$
to the 2D photo $F$ by some distance $D(F,A M_\theta+b)$,
e.g.\ square or absolute or Mahalanobis.
We do not want to assume any extra knowledge like the lighting
conditions $A$ under which the photo has been taken,
which rules out a direct use of $D$. Ideally we want
a ``distance'' between $F$ and $M$ that is independent of $A$
and is zero if and only if there exists a lighting condition
$A$ such that $F$ and $AM_\theta+b$ coincide.

Indeed, this is possible, if (rather than defining $M_\theta$ as some
$A$-dependent rendered projection of $M$) we use $A$-independent
brightness and normals $M_\theta$ as pixel features as defined above,
and define a linearly invariant distance as follows. Let

\bqa
  \bar F    &:=& \textstyle {1\over |P|}\sum_{p\in P} F(p) \;\;\; \in\SetR  \nonumber\\
  \bar M_\theta &:=& \textstyle {1\over |P|}\sum_{p\in P}  M_\theta(p) \;\in\SetR^4
\eqa

be the average attribute values of photo and projection, and
\beq
  C_{F M_\theta} := {1\over |P|}\sum_{p\in P}(F(p)-\bar F)(M_\theta(p)-\bar M_\theta)^\trp\;\in\SetR^{1\times 4}
\eeq
be the cross-covariance matrix between $F$ and $M_\theta$ and similarly
$C_{M_\theta F}=C_{FM_\theta }^\trp\in\SetR^{4\times 1}$ and the covariance matrices
$C_{FF}\in\SetR^{1\times 1}$ and $C_{M_\theta M_\theta}\in\SetR^{4\times 4}$.
Consider the following  distance or loss function between $F$ and $M_\theta$ which is
obtained from \req{defLILGeneral} derived in Section \ref{secLossFunction}, when $X=F$, $Y=M_\theta$ $n=1$ and $m=4$.
\beq\label{defLIL}
\Loss(\X) := 1 - \tr[C_{FM_\theta}C_{M_\theta M_\theta}^{-1}C_{M_\theta F}C_{FF}^{-1}]
\eeq
Obviously this expression is independent of $A$. In the next
section we show that it is invariant under regular linear
transformations of the pixel/attribute values of $F$ and $M_\theta$
and zero if and only if there is a perfect linear transformation of
the pixel values from $M_\theta$ to $F$.
This makes it unnecessary to know the exact surface reflection constants of the object ($k_a(p)\in\SetR$ and $k_d(p)\in\SetR$).
We will actually derive
\beq
  \Loss(\X) = \smash{\min_{A,b}}\,D_{\text{Mahalanobis}}(F,A\!\cdot\!M_\theta+b)
\eeq
This implies that $\Loss(\X)$ is zero if and only if there
is a lighting $A$ under which $F$ and $M_\theta$ coincide, which we
desired.

%------------------------------------------------------------------------
\section{Derivation of the Invariant Loss Function}\label{secLossFunction}
%------------------------------------------------------------------------
A detailed derivation of the loss function is given in this section.

\paradot{Notation}
Using the notation of the previous section, we measure the
similarity of photo $F:P\to\SetR^n$ and projected 3D model
$M_\theta:P\to\SetR^m$ (returning to general $n,m\in\SetN$) by some
loss:
\beq\label{defLoss}
  \Loss(\X) := D(F,M_\theta)
  := {1\over|P|}\smash{\sum_{p\in P}} d(F(p),M_\theta(p))
\eeq
where $d$ is a distance measure between corresponding pixels of the
two images to be determined below. A very simple, but as discussed
in Section \ref{secRelatedWork} for our purpose unsuitable,
choice in case of $m=n$ would be the square loss
$d(F(p),M_\theta(p))=||F(p)-M_\theta(p)||_2^2$.

It is convenient to introduce the following probability notation:
Let $\omega$ be uniformly distributed\footnote{With a non-uniform distribution
one can easily weight different pixels differently.} in $P$, i.e.\
$\P[\omega]=|P|^{-1}$. Define the vector random variables
$X:=F(\omega)\in\SetR^n$ and $Y:=M_\theta(\omega)\in\SetR^m$.
The expectation of a function of $X$ and $Y$ then is
\beq
  \E[g(X,Y)]:={1\over|P|}\sum_{\omega\in P}g(X(\omega),Y(\omega))
\eeq
With this notation, \req{defLoss} can be written as
\beq
  \Loss(\X) = D(X,Y) = \E[d(X,Y)]
\eeq

\paradot{Noisy (un)known relation}
Let us now assume that there is some (noisy) relation $f$ between
(the pixels of) $F$ and $M_\theta$, i.e.\ between $X$ and $Y$:
\beq
  Y=f(X)+\eps,\qquad \eps=\mbox{noise}
\eeq
If $f$ is known and $\eps$ is Gaussian, then
\beq
  D_f(X,Y)=\E[||f(X)-Y||_2^2]
\eeq
is an appropriate distance measure for many purposes.
In case $F$ and $M_\theta$ are from the same source (same pixel attributes, lighting
conditions, etc), choosing $f$ as the identity function results in a standard square loss.
In many practical applications, $f$ is not the
identity and furthermore unknown (e.g.\ mapping gray models to real
color photos of unknown lighting condition). Let us assume $f$
belongs to some set of functions $\cal F$. $\cal F$ could be the set
of all functions or just contain the identity or anything in between
these two extremes. Then the ``true/best'' $f$ may be estimated by
minimizing $D_f$ and substituting into $D_f$:

\bqa
  f_{best} &=& \arg\min\nolimits_{f\in\cal F} D_f(X,Y) \\
  D(X,Y)  &:=& \min\nolimits_{f\in\cal F}D_f(X,Y)
\eqa

Given $\cal F$, $D$ can in principle be computed and
measures the similarity between $X$ and $Y$ for unknown $f$.
Furthermore, $D$ is invariant under any transformation $X\to g(X)$
for which ${\cal F}\circ g=\cal F$.

\paradot{Linear relation}
In the following we will consider the set of linear relations
\beq
  {\cal F}_{lin} \;:=\; \{f:f(X)=AX+b,A\in\SetR^{m\times n},b\in\SetR^m\}
\eeq
For instance, a linear model is appropriate for mapping color to
gray images (same lighting), or positives to negatives.
For linear $f$, $D$ becomes
\beq
  D(X,Y) \;=\; \min_{A\in\SetR^{m\times n}}\min_{b\in\SetR^m}
               \E[||AX+b-Y||_2^2]
\eeq
and the distance is invariant under all regular
linear re-parametrization of $X$, i.e.\ $D(X,Y)=D(AX+b,Y)$ for all
$b$ and all non-singular $A$. Unfortunately, $D$ is not symmetric in
$X$ and $Y$ and in particular not invariant under linear
transformations in $Y$.
Assume that the components $(Y_1,...,Y_m)^\trp$ are very different to each other ($Y_1$=color, $Y_2$=angle, $Y_3$=texture).
Then the 2-norm $||Y||_2^2=Y^\trp Y=Y_1^2+...+Y_m^2$ does not take these differences into account.
A standard solution is to normalize
by the variance, i.e.\ use $\sum_i Y_i^2/\sigma_i^2$, where
$\sigma_i^2=\E[Y_i^2]-\E[Y_i]^2$, but this norm is (only) invariant
under component scaling.

\paradot{Linearly invariant distance}
To get invariance under general linear transformations, we have to
scale by the covariance matrix
\beq
  C_{YY} \;:=\; \E[(Y-\bar Y)(Y-\bar Y)^\trp],\quad \bar Y:=\E[Y]
\eeq
The Mahalanobis norm (cf.\ Section \ref{secRelatedWork})
\beq
  ||Y||_{C_{YY}^{-1}}^2 \;:=\; Y^\trp C_{YY}^{-1} Y
\eeq
is invariant under linear {\em homogeneous} transformations, as can be seen from
\bqa
  ||AY||_{C_{AY,AY}^{-1}}^2
  &\equiv & Y^\trp A^\trp C_{AY,AY}^{-1}AY \nonumber\\
  &=& Y^\trp C_{YY}^{-1} Y				\nonumber\\
  &\equiv & ||Y||_{C_{YY}^{-1}}^2
\eqa
where we have used $C_{AY,AY}=A C_{YY} A^\trp$.

The following distance is hence invariant under {\em any}
non-singular linear transformation of $X$ and any non-singular
(incl.\ non-homogeneous) linear transformation of $Y$:
\beq\label{defDinv}
  D(X,Y) \;:=\; \min_{A\in\SetR^{m\times n}}\min_{b\in\SetR^m}
               \E[||AX+b-Y||_{C_{YY}^{-1}}^2]
\eeq

\paradot{Explicit expression}
Since \req{defDinv} is quadratic in $A$ and $b$, the minimization can
be performed explicitly, yielding  after some linear algebra 
\bqa\label{Abmin}
  b  & = & b_{min} := \bar Y-A_{min}\bar X, \quad \bar X:=\E[X]\quad \nonumber\\
  A  & = & A_{min}  := C_{YX}C_{XX}^{-1}
\eqa
where $C_{XY} =  \mbox{Cov}(X,Y) = \E[(X-\bar X)(Y-\bar Y)^\trp]$
and similarly $C_{YX}=C_{XY}^\trp$ and $C_{XX}$. Inserting
\req{Abmin} back into \req{defDinv} and rearranging terms gives
\bqa
 D(X,Y) &=& \tr[1\!\!1-C_{YX}C_{XX}^{-1}C_{XY}C_{YY}^{-1}] \nonumber \\
        &=&   m - \tr[C_{XY}C_{YY}^{-1}C_{YX}C_{XX}^{-1}]
\eqa
This explicit expression shows that
$D$ is symmetric in $X$ and $Y$ if not for the $m$ term.
For comparisons, e.g.\
for minimizing $D$ w.r.t.\ $\theta$, the constant $m$ does not matter.
Since the trace can assume all and only values in the interval
$[0,\min\{n,m\}]$, it is natural to symmetrize $D$ to obtain
  \bqa  
    \Loss(X,Y) &=& \min\{D(X,Y),D(Y,X)\} \nonumber \\
    &=& \min\{n,m\}-\tr[C_{XY}C_{YY}^{-1}C_{YX}C_{XX}^{-1}]
    \label{defLILGeneral}
  \eqa
Returning to original notation, this expression coincides with the
loss \req{defLIL}. It is hard to visualize this loss, even for $n=1$
and $m=4$, but the special case $m=n=1$ is instructive, for which
the expression reduces to
$D(X,Y)=1-\mbox{corr}^2(X,Y)$, where
$\mbox{corr}(X,Y) = {\mbox{Cov}(X,Y)/\sigma_X\sigma_Y}$
is the correlation between $X$ and $Y$. The larger the (positive or
{\em negative}) correlation, the more similar the images and the
smaller the loss. For instance, a photo has maximal correlation with
its negative.

%----------------------------------------------------------------------------
\section{Practical Behaviour of the Loss Function}\label{secLossLandscape}
%----------------------------------------------------------------------------
In this section, we explore the  nature of the loss function derived
in Section~\ref{secLossFunction} for real and synthetic
photographs.

\begin{figure}[t!]
\centering
  \includegraphics[trim = 0px 20px 0px 20px, clip, width=0.48\textwidth]{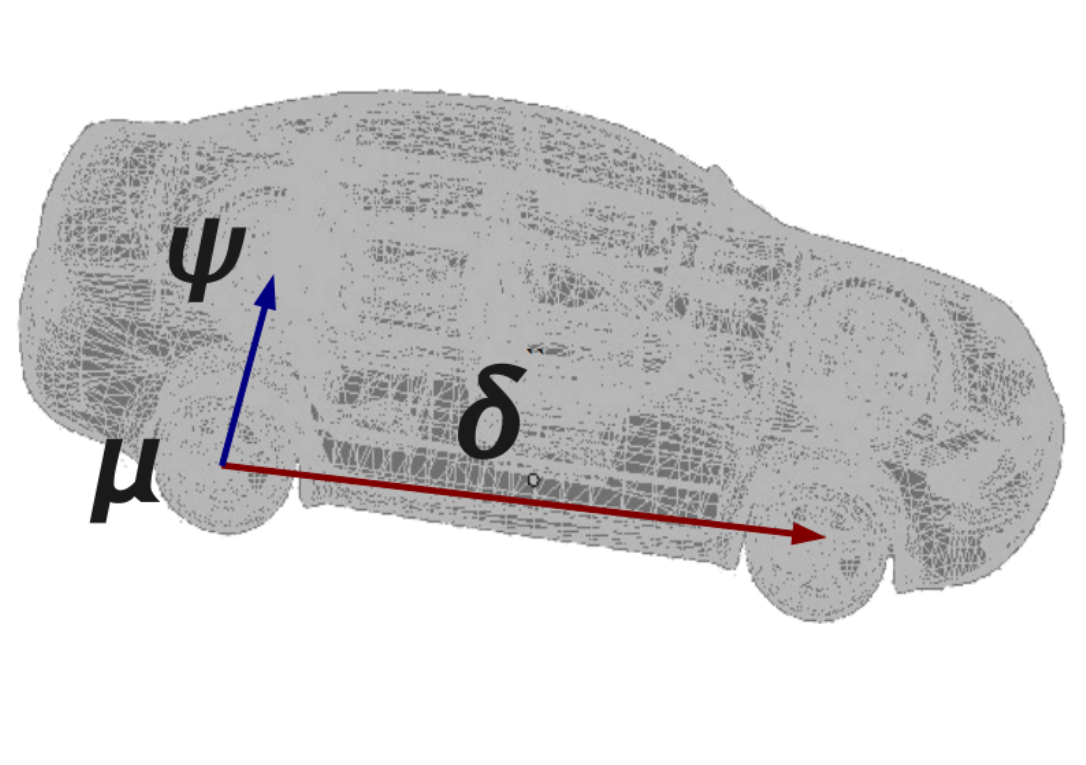}
\caption[Pose representation] {
The pose representation $\Xortho$
used for 3D car models. We use the rear wheel center $\v\mu$,
the vector between the wheel centers $\v\delta$ and unit vector $\v\psi$ in
the direction of the rear wheel axle.}
\label{fig:poseRepresentationX}
\end{figure}

\paradot{Representation of the pose}
The pose of a generic object may be represented by translations along the X,Y and Z axes and a suitable rotation representation. A quaternion or exponential map \cite{Lepetit05monocularmodel-based} based rotation can be used to avoid \emph{Gimbal Lock} problems that may occur with Euler angles or roll/yaw/pitch based rotations.
Careful selection of pose parameters can aid the optimizer when finding the best pose.
Since we work with vehicles,  the following pose representation was used, temporarily neglecting the effects of perspective projection.
It is consistent with the  rough pose estimation method described in~\cite{hutter2009matching}.
\beq
\label{eqn:PoseX}
\Xortho := \left( \mu_x, \mu_y, \delta_x, \delta_y, \psi_x, \psi_y \right)
\eeq
$\v\mu=(\mu_x, \mu_y)$ is the visible rear wheel center of the
car in the 2D projection. $\v\delta=(\delta_x, \delta_y)$ is
the vector between corresponding rear and front wheel centers
of the car in the 2D projection.
The 2D image is a projection of the 3D model on to the XY
plane.
$\v\psi = \left( \psi_x, \psi_y, \psi_z \right)$ is a unit
vector in the direction of the rear wheel axle of the 3D car
model. Therefore, $\psi_z = -\sqrt{1 - \smash{\psi_x^2 -
\psi_y^2}}$ and need not be explicitly included in the pose
representation $\X$. This representation is illustrated in
Figure ~\ref{fig:poseRepresentationX}.
This pose is converted to OpenGL translation, scale and rotation as per
\cite{hutter2009matching} to transform and project the 3D model. As we directly optimize \req{defLIL} w.r.t $\X$ (Section \ref{secOptim}) explicit knowledge of intrinsic camera parameters etc., are not required.

\begin{figure}[htp]
\subfigure[Perspective projection model]{\label{fig:perspProjModel}
	\includegraphics[width=0.48\textwidth]{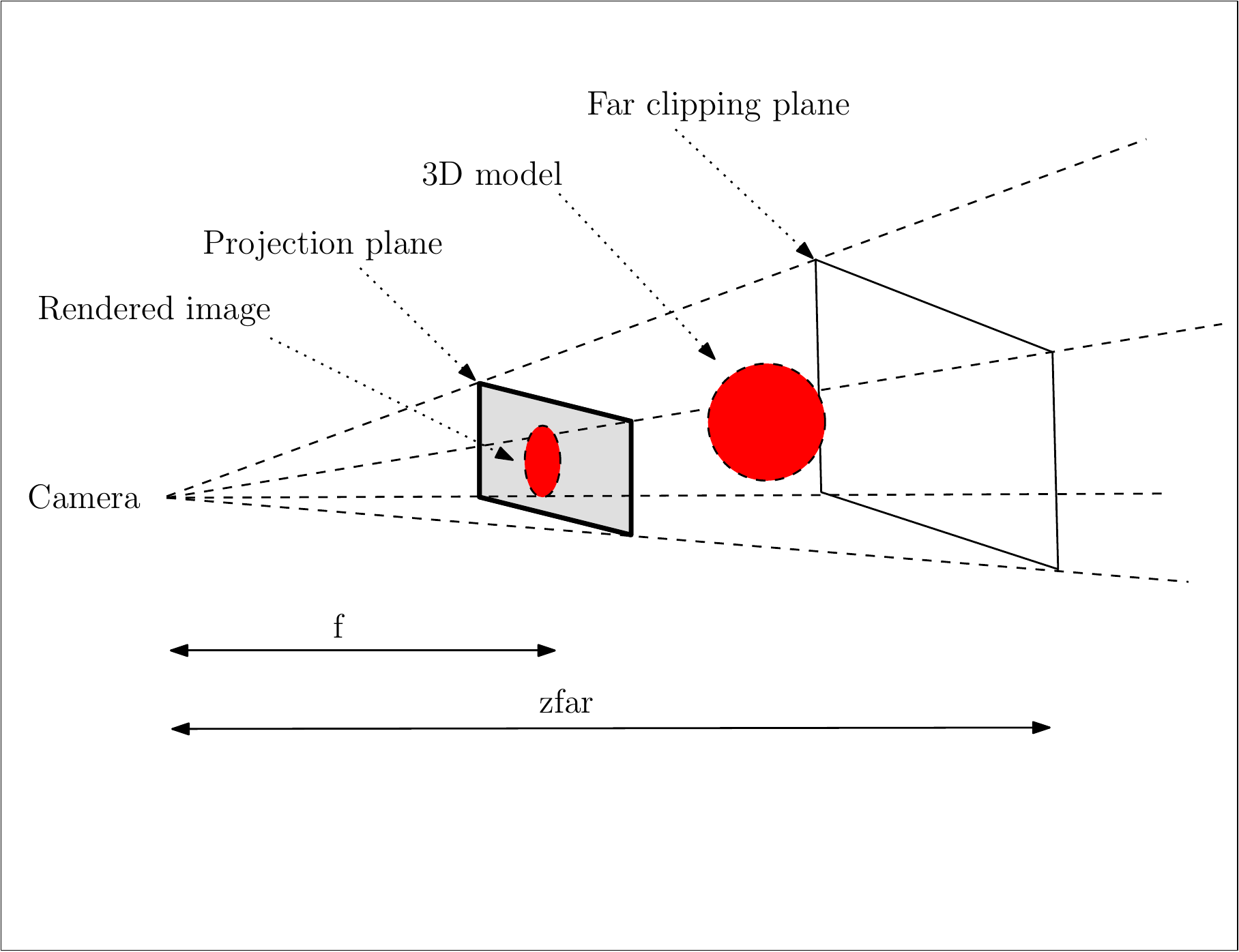}}
\subfigure[Handling object clipping]{\label{fig:perspProjClipping}
	\includegraphics[width=0.48\textwidth]{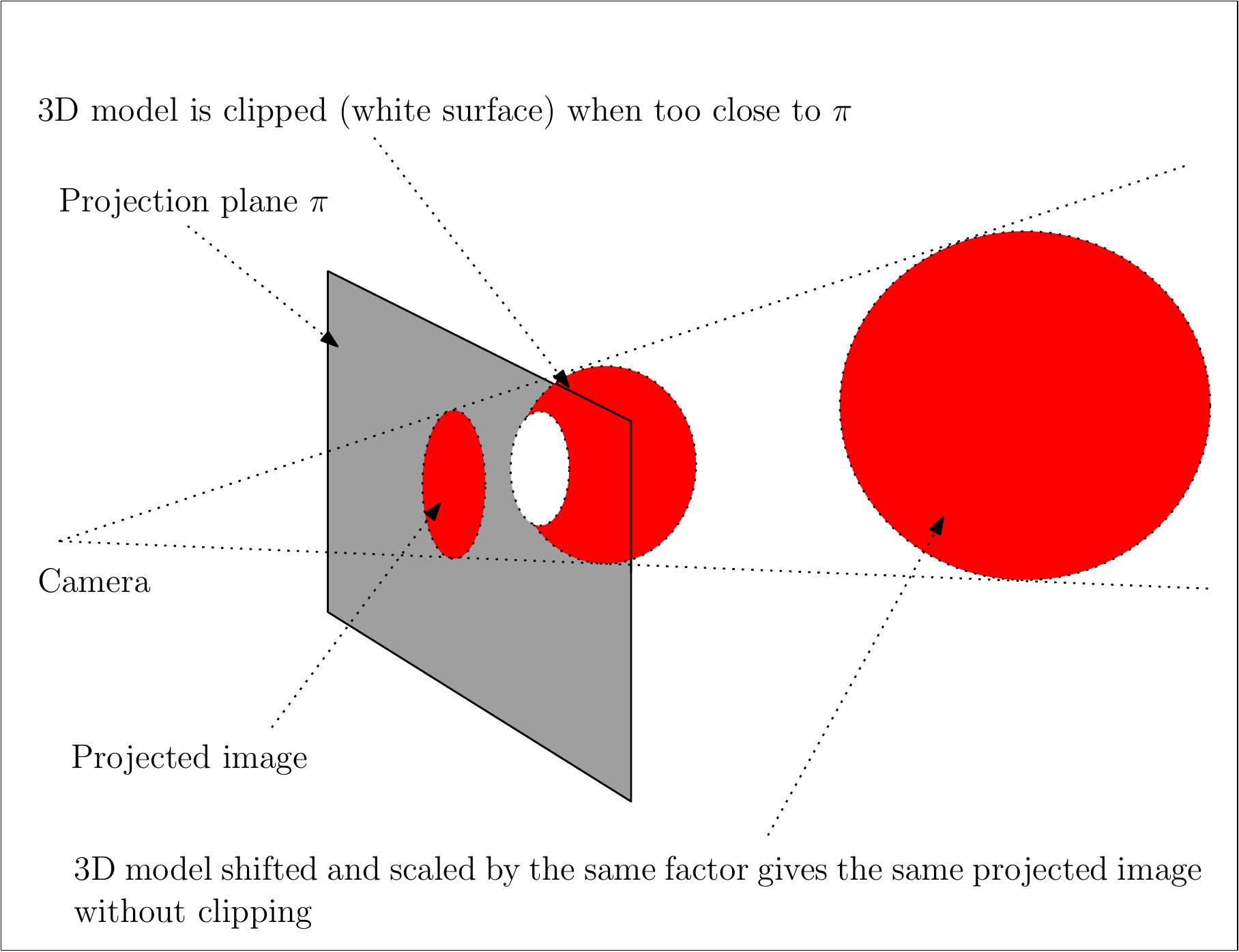}}
\caption[Rendering with perspective projection]
{Rendering with perspective projection. \ref{fig:perspProjModel} shows the perspective projection model used.
\ref{fig:perspProjClipping} illustrates clipping when the 3D model is located too close to the projection plane $\pi$ and how this is prevented.}
\end{figure}

\paradot{Perspective projection}
The pose estimation was extended to handle perspective projection as follows.
The 3D model was rendered using the OpenGL perspective projection model.
The degree of perspective distortion was changed by varying the parameter $f$ (Figure \ref{fig:perspProjModel}) in the OpenGL frustum. $f$ was included as a pose parameter during the optimization. The 3D model is sometimes clipped by the projection plane $\pi$ when positioned too close to $\pi$. We avoid this by shifting and scaling the 3D model  by a constant factor $\alpha$ (Figure \ref{fig:perspProjClipping}), thus obtaining the same projected image without clipping.

Thus the parallel projection based pose in \req{eqn:PoseX} becomes.
\beq
\label{eqn:PoseXpersp}
\Xpersp := \left( \mu_x, \mu_y, \delta_x, \delta_y, \psi_x, \psi_y, f \right)
\eeq

\begin{figure}[t!]
\centering\hspace{-3ex}
\def\w{0.48}
\subfigure[Synthetic photo]{\label{fig:F1Surf}
	 \includegraphics[width=\w\textwidth]{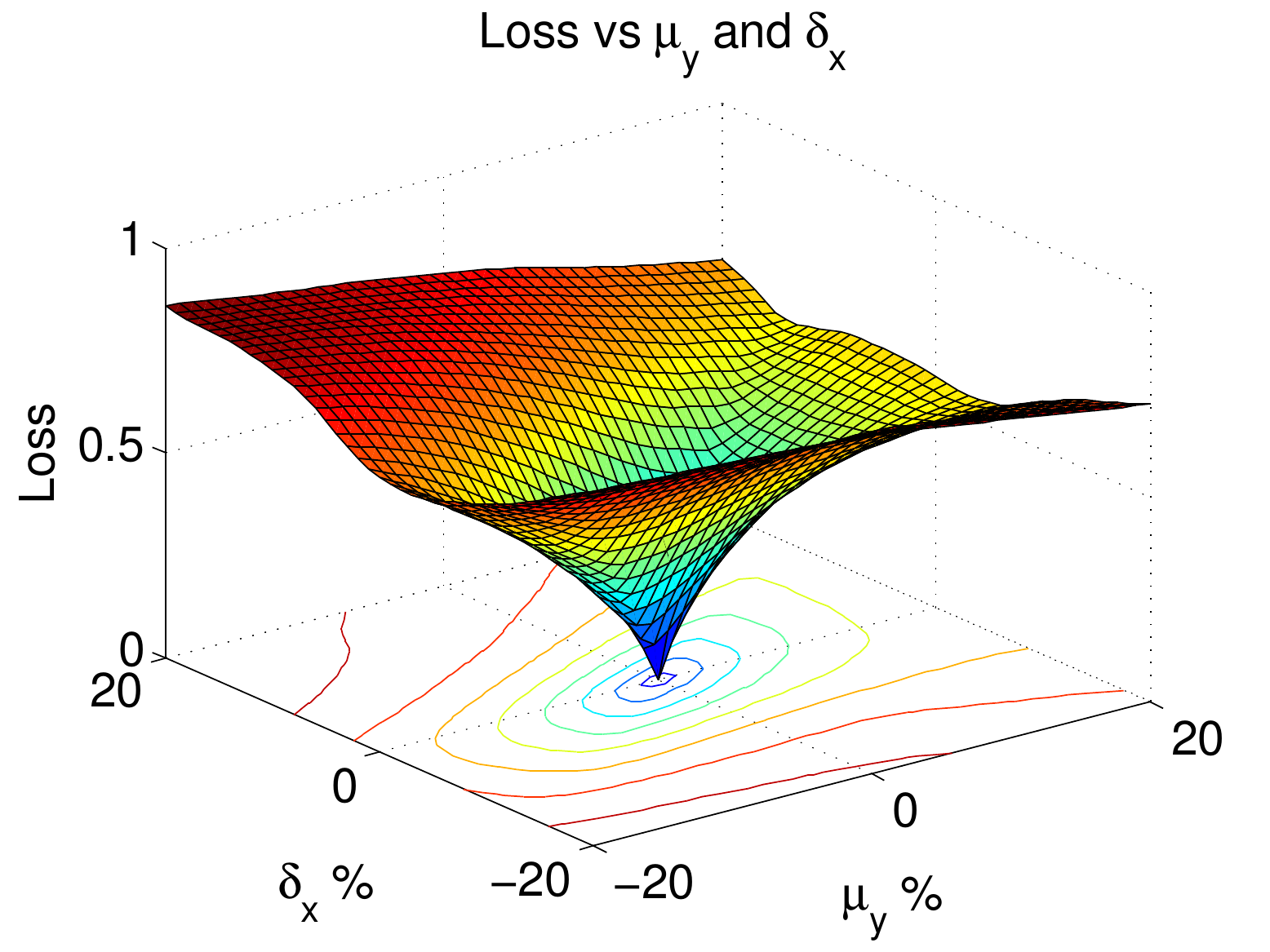}}
\hspace{-2ex}
\subfigure[Real photo]{\label{fig:LosslandscapeRealPhoto}
	\includegraphics[width=\w\textwidth]{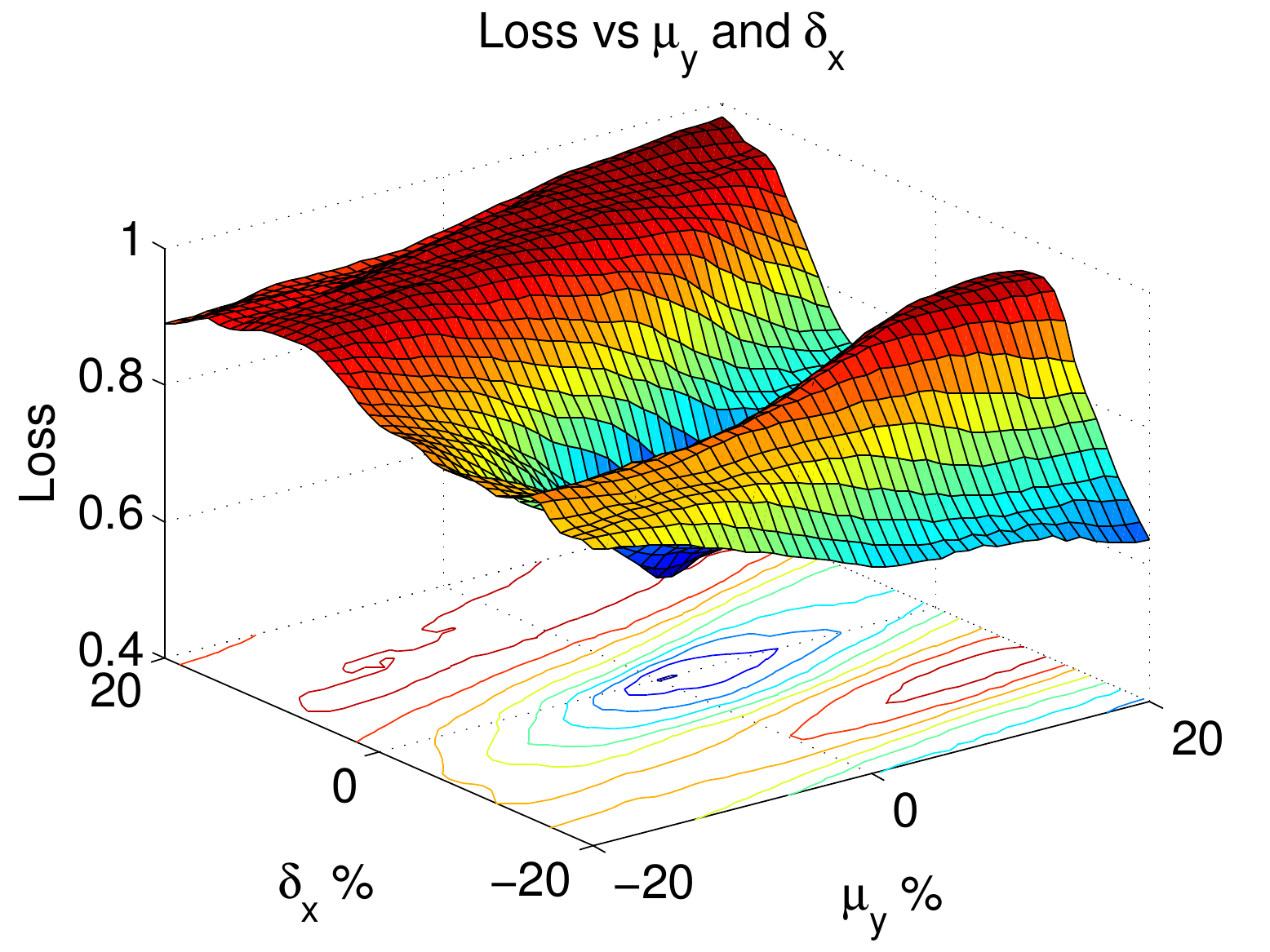}}
\hspace{-1ex}
\caption[Loss landscapes for synthetic and real photos]{ {\bf Loss
landscapes for synthetic and real photos.} The six dimensional loss
function was visualized by plotting its variation with a pair of
pose parameters at a time. Based on our pose representation this
results in fifteen plots. The variation of the loss function with  a
pair of pose parameters are shown  for a synthetic photograph and
a real photograph. The nature of the loss function for real
photographs makes it more difficult to find the global minimum
(hence the correct pose) than for synthetic photographs.}
\label{fig:LossLandscapeArtificialAndRealPhotos}
\end{figure}

\paradot{Loss landscape for synthetic photographs}
To understand the behavior of the loss function, we have generated
loss landscapes for synthetic images of 3D  models. To produce
these landscapes, a synthetic photograph was generated by
projecting the 3D model at a known pose $\X_0$ with Phong shading.
We then vary the pose parameters, two at a time about $\X_0$ and
find the value of the loss function between this altered projection
and the ``photograph'' taken at $\X_0$. These loss values are recorded,
allowing us to visualize the behavior of the loss function by
observing surface and contour plots of these values. The unaltered
pose values should project an image identical to the input
photograph, giving a loss of zero according to the loss function
derived in Section~\ref{secLossFunction}, with a higher loss
exhibited at other poses.
The variation of the loss with respect to a pair of pose parameters
is shown in Figure \ref{fig:F1Surf}. It can be seen from these loss
landscapes that the  loss has a clear minimum at the initial pose
$\X_0$. The loss values increase as these pose parameters deviate
away from $\X_0$, up to $\pm 20\%$.
From this data, we are able to see that the minimum corresponding to
$\X_0$ can be considered a global minimum for all practical purposes.
The shape of the surface plots was similar for all other parameter pairs,
indicating that the
complete landscape of the loss function should
similarly have a global minimum at the initial pose, allowing us to
find this point using standard optimization techniques, as
demonstrated in Section~\ref{secOptim}.

\paradot{Loss landscape for real photographs}
The landscape of the loss function was analyzed for real photographs
by varying the pose parameters of the model about a pose obtained by
manually matching the 3D car model to a real photograph. The variation was plotted by taking a
pair of pose parameters at a time over the entire set of pose
parameters. A loss landscape obtained by varying $\mu_y$ and
$\delta_x$ for a real photograph is shown in Figure
\ref{fig:LosslandscapeRealPhoto}. The variation of the loss function
for other pose parameter pairs were found to be similar. Although a
global minimum exists at the best pose of the real photograph, the
nature of the loss function surface makes it more difficult to
optimize when compared to synthetic photos (Figure
\ref{fig:F1Surf}). In particular, one can observe local minima and the
landscape in higher dimensions is
considerably more complex.

%------------------------------------------------------------------------
\section{Optimizing the  Loss Function for Pose Estimation}\label{secOptim}
%------------------------------------------------------------------------
As explained in Section~\ref{secLossFunction}, the correct pose
parameters $\v{\theta_{opt}}$ will  give  the lowest loss value. The
loss function landscape, as discussed in Section~\ref{secLossLandscape}, shows that $\v{\theta_{opt}}$ corresponds to the
global minimum of the loss function.
Therefore, the loss function \req{defLIL} was minimised w.r.t. $\X$ to obtain $\v{\theta_{opt}}$.
The optimization strategy is described in this section.

\paradot{The optimizer}
To immunise the optimisation from pixel quantisation artefacts and
noise in the images, direct search methods that do not calculate the
derivative of the loss function were considered.
The optimization was performed using the well known \emph{Downhill Simplex
Method (DS)}  \cite{nelder1965simplex,nr2,matlab}, owing to its efficiency and
robustness.
When optimizing an $n$-dimensional function with the DS method, a so
called \emph{simplex} consisting of $n+1$ points is used to traverse
the $n$-dimensional search space and find the optimum.

The reliability of the optimization is adversely affected by the
existence of local minima. Fortunately, the \emph{Downhill Simplex}
method has a useful property. In most cases, if the simplex is
reinitialized at the pose parameters of the local minimum and the
optimization is performed again, the solution converges to the
global minimum.
Proper parameterization is important for the optimizer to give good
results. We have used a normalized pose parameterization as follows.

\paradot{Normalised pose parameters}
Normalization gives each pose parameter a comparable range during
optimization.
The normalized pose $\v{\X_N}$ was obtained by normalizing
the pose  w.r.t.\ the dimensions of the photograph as follows.
\beq
 \v{\X_N} = \left( \frac{\mu_x}{I_W} , \frac{\mu_y}{ I_H} , \frac{\delta_x}{ I_W}, \frac{\delta_y }{I_H}  , \psi_{x}, \psi_{y}, \frac{10I_W}{f} \right)
\eeq
$ I_W , I_H $ are the width and height of the photograph (2D image).
$\v\psi$ is a unit vector and does not require normalization.

\paradot{Initialisation}
The downhill simplex method, like all optimization techniques,
requires a reasonable starting position. There are many methods for
selecting a starting point, from repeated random initialization to
structured partitioning of the optimization volume. A disadvantage
of these methods is that they require a number of optimization runs
to locate the optimal point, which can take significant time.
Depending on the application, it may be possible to develop a coarse
location method which provides an estimate of the initial pose.
Possible methods for obtaining a coarse initial pose include the work done by \cite{OzuysalLF09}, \cite{depthEncodedHoughVotingEccv2010} and \cite{implicitshapepose}.
We have used the wheel match method described in~\cite{hutter2009matching}.
to obtain an initial pose for vehicle photos where the wheels are visible. The wheels need not be visible with the other methods mentioned above.
Since the wheel match method gives the pose for an orthogonal projection, the perspective parameter $f$ was initialized to a large value as $f=10I_w$ to get negligible perspective distortion in the initial rough pose used for the optimization.

\paradot{Background removal}\label{ClipBG}
As the effects of the background clutter in the photo adds considerable noise to the loss function landscape we use an adaptation of grabcut \cite{grabcut} to remove a considerable amount of the background pixels from the photo. The initial rough pose estimate is used as a prior to generate the background and foreground grabcut masks \footnote{We use the cv::grabCut() method provided in OpenCV \cite{opencv_library} version 2.1}. The masks are obtained by scaling the model projection obtained from the initial pose by a margin $m$.

%------------------------------------------------------------------------------
\section{Experimental Results}\label{secResults}

\begin{figure*}[t!]
\centering
\def\w{0.23}
\subfigure[Hyundei Getz]{
     \includegraphics[width=\w\textwidth]{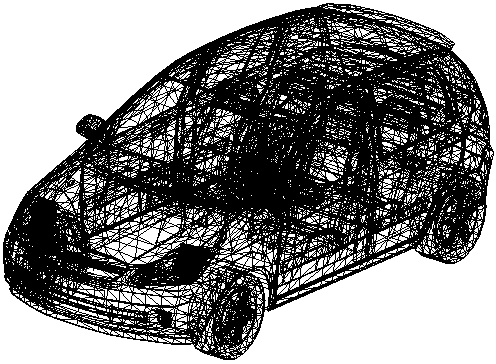}}
\subfigure[Jeep Cherokee]{
     \includegraphics[width=\w\textwidth]{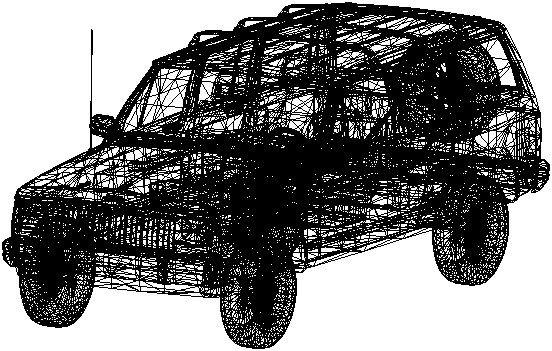}}
\subfigure[Mazda 3]{
     \includegraphics[width=\w\textwidth]{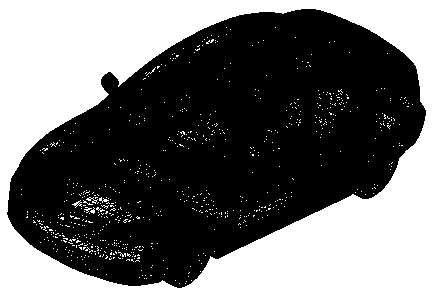}}
\subfigure[Mazda Astina (scanned)]{\label{fig:modelMazdaAstina}
     \includegraphics[width=\w\textwidth]{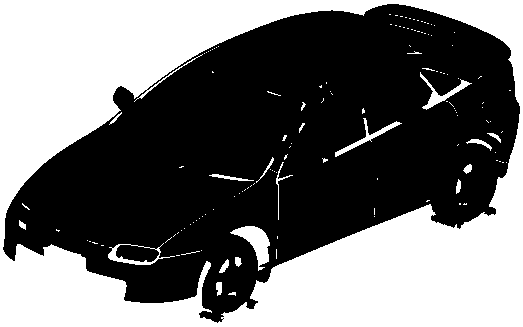}}
\caption[3D CAD Models]{Some of the 3D CAD models used for the experiments are shown. \ref{fig:modelMazdaAstina} is a laser scanned 3D model of a real Mazda Astina car and matches the proportions and detail of the real car in Figure \ref{fig:ResultsAstinaBest}. Additionally it has a very high number of polygons.}
\label{fig:3DModels}
\end{figure*}

\begin{figure*}[t!]
\centering
\def\w{0.48}
\def\tp{T269_Hyundai-Getz-Rear-View_jpgTweak_detect_ellipse_}
\subfigure[Initial]{
	 \includegraphics[width=\w\textwidth]{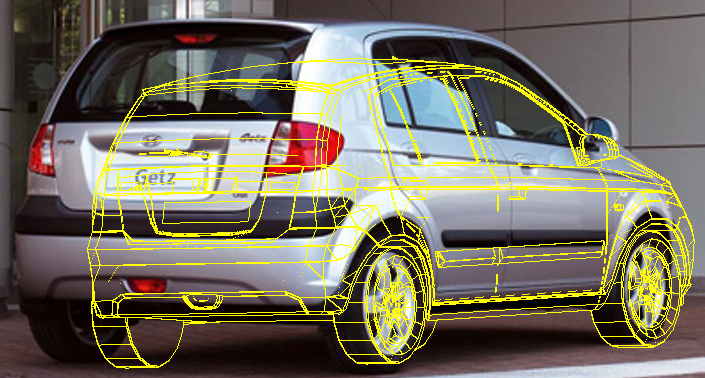}}
\subfigure[Final]{
	 \includegraphics[width=\w\textwidth]{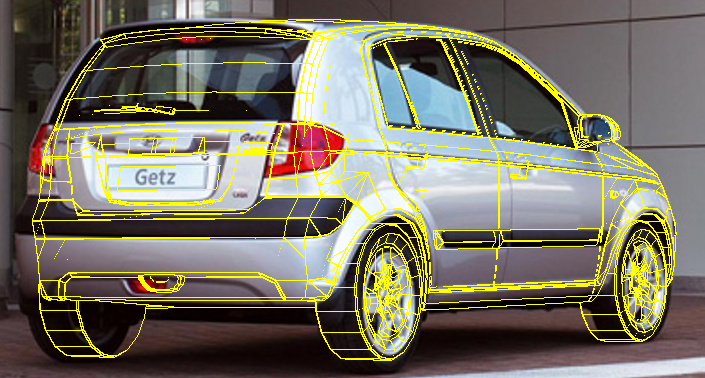}}
\def\tp{T269_DSC01647_LR800x600_}
\subfigure[Initial]{
	 \includegraphics[width=\w\textwidth]{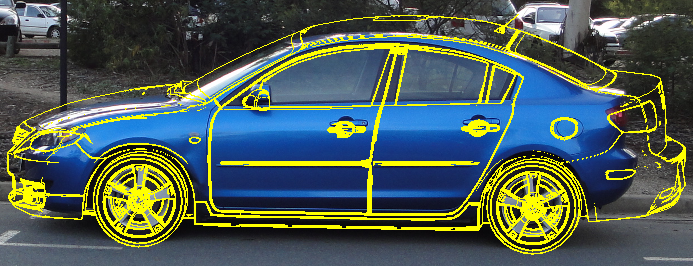}}
\subfigure[Final]{
	 \includegraphics[width=\w\textwidth]{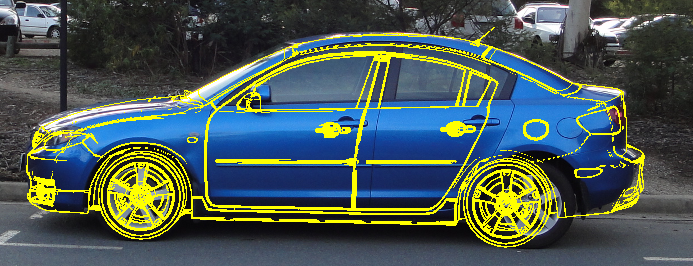}}
\def\tp{jeep1gcbb_}
\subfigure[Initial]{
	 \includegraphics[width=\w\textwidth]{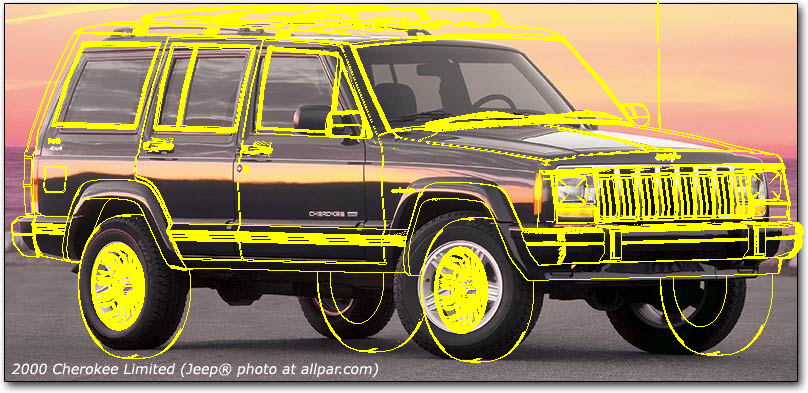}}
\subfigure[Final]{
	 \includegraphics[width=\w\textwidth]{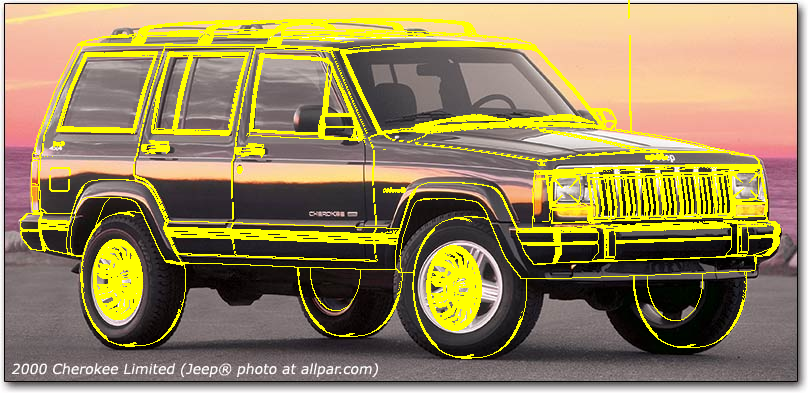}}
\def\tp{T269_IMG_3019800x533_}
\subfigure[Initial]{
	 \includegraphics[width=\w\textwidth]{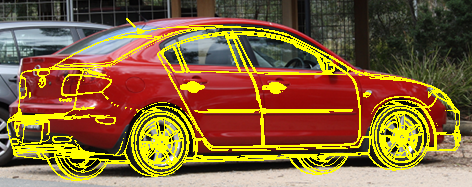}}
\subfigure[Final]{
	 \includegraphics[width=\w\textwidth]{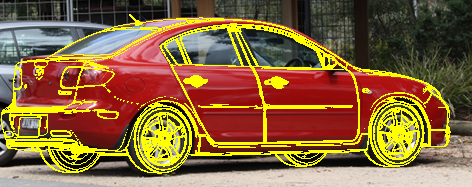}}
\caption[Results]{ {\bf Experimental Results} The `Initial rough pose' (column 1) used to initialize the optimization and the resulting  `Final pose' (column 2) obtained by optimizing the novel loss function are shown for different real photos (row-wise). The pose is shown in `yellow' by an outline of the projected 3D model. Unlike the scanned 3D model in Figure \ref{fig:ResultsAstinaBest}, these 3D models do not perfectly match the proportions and detail of the real vehicle in the photo. However, the proposed method produces good results even with approximate 3D models. The images have been cropped for visual clarity.}
\label{fig:ResultsGroup}
\end{figure*}

Experimental results on real photos of different vehicle types and colours (using corresponding 3D Models) are shown in Figures \ref{fig:ResultsAstinaBest} and \ref{fig:ResultsGroup}. The photos have realistic conditions like cast shadows and surface specularities. We have used a laser scanned 3D CAD model of a Mazda Astina (with more than 2 million polygons), a Mazda~3 3D model, a Jeep Cherokee 3D model and a Hyundai Getz 3D model (Figure~\ref{fig:3DModels}). The latter models were obtained from the Internet and have  less than 500,000 polygons. The optimization was done using perspective projection. A perfect 3D pose is recovered with the scanned 3D model (Figure \ref{fig:ResultsAstinaBest}). The 3D models obtained from the Internet do not match the proportions and details of the real vehicles exactly. However, we see good results even when the 3D models do not perfectly represent the object in the scene (Figure \ref{fig:ResultsGroup}).

%------------------------------------------------------------------------
\section{Implementation Details}\label{secTech}
%------------------------------------------------------------------------
In this section we describe some of the technical  aspects of the
proposed work.  The initial code  was implemented in MATLAB \cite{matlab}, however, components were gradually ported to C in
order to improve performance.

\paradot{3D rendering}
In order to calculate the loss values described in
Section~\ref{secLossFunction}, it was required to render the surface
normals and brightness of a 3D model at a given pose. Initially, the rendering was
done using \emph{model3D} \cite{model3d}, a BSD licensed MATLAB
 \cite{matlab} class. As this rendering was not fast enough for our
application, a separate module was written in C to render the model
off-screen using OpenGL \cite{opengl}
 pBuffer extension and GLX.
This C module was  used with the MATLAB code using the \emph{MEX} gateway.
 Initially, only the rendering was done in C.
The rendered 2D intensity and surface normal matrices were returned
back to MATLAB using the MEX gateway.
 This seemed to exhaust memory
during the reliability tests described in Section~\ref{secOptim}.
 Therefore, the r endering and the loss calculation were also implemented
in C, with only the loss value returned to MATLAB for use in optimization.
\begin{table}
  \caption{Rendering and loss calculation times.}
  \label{tblRenderLosscalcTime}
  \begin{center}
    \begin{tabular}{lcr}
  \hline\noalign{\smallskip}
  \bf Approach   & \bf Loss calc.& \bf Render\\
  \noalign{\smallskip}\hline\noalign{\smallskip}
  MATLAB     & 0.16 s     & 2.28 s \\
  C/OpenGL   & 0.04 s     & 0.17 s \\
  \noalign{\smallskip}\hline
 \end{tabular}
 \end{center}
\end{table}
This second approach improved performance in terms of speed and memory
usage. A summary of the time taken to render the image and to
calculate the loss using these approaches are presented in
Table \ref{tblRenderLosscalcTime}.

\paradot{Running times}
A typical Downhill Simplex minimization required in the order of
100--200 loss function evaluations.
Using the C based loss calculation and OpenGL rendering, pose
estimation in synthetic images took around  1 minute for models with more than 30,000 nodes.
Recent work done in \cite{moreno2008pose} on pose estimation using
point correspondences, takes more than 3 minutes (200 seconds) for
a synthetic image of a model with only 80 points.
Hence, despite being a pixel based method, the performance of our
approach is very encouraging.
Further improvements in speed may be obtained by using the graphics hardware (GPU) for computing the loss function.

%----------------------------------------------------------------
\section{Discussion}\label{secDisc}
%----------------------------------------------------------------
A method to register a known 3D model on a given 2D image is
presented in this paper.
A \emph{novel distance measure} ({\sec} \ref{secLossFunction}) between attributes in the
2D image and projected 3D model is optimized to recover the 3D pose of the object in the given image.
Pose estimation results on real photos of different vehicles are shown with the  optimization initialized from a rough pose obtained using wheel locations~\cite{hutter2009matching}.
The method differs from existing 2D-3D
registration methods found in the literature.
The proposed method requires only a single view of the object. It
does not require a motion sequence and works on a static image from
a given view. Also, the method does not require the camera
parameters to be known a priori. Explicit point correspondences or
matched features (which are hard to obtain when comparing 3D models
and image modalities) need not  be known before hand. The method can
recover the full 3D pose of an object. It does not require prior
training or learning. As the method can handle 3D models of high
complexity and detail, it could be used for applications that
require detailed analysis of 2D images. It is particularly useful in
situations where a known 3D model is used as a ground truth for
analyzing a 2D photograph.
The method has been currently tested on real and synthetic
photographs of cars with promising results.

\paradot{Outlook}
A planned application of the method is to analyze images of damaged
cars. A known 3D model of the damaged car will be registered on the
image to be analyzed, using the proposed registration method. This
will be used as a ground truth. The method could be extended further
to simultaneously identify the type of the car while estimating its
pose, by optimizing the loss function for a number of 3D models and
selecting the model with the lowest loss value.
More sophisticated optimization methods may be used to improve results further.

\paradot{Conclusion}
We conclude from our results that the linearly invariant loss
function derived in Section~\ref{secLossFunction} can be used to
estimate the pose of cars from real photographs. We also demonstrate
that the \emph{Downhill Simlpex} method can be effectively used to
optimize the loss function in order to obtain the correct pose.
Allowing simplex re-initializations makes the method more robust
against local minima.
Despite being a direct pixel based method (as opposed to a feature/point based
method), the performance of our method is very encouraging in
comparison with other recent approaches, as discussed in
Section~\ref{secTech}.

\paradot{Acknowledgment}
This work was supported by Control\Euro xpert.
The authors wish to thank Stephen Gould and
Hongdong Li for the valuable feedback and advice.

\bibliographystyle{spmpsci}      

\end{document}